\documentclass[10pt,twocolumn,letterpaper]{article}

\usepackage{iccv}
\usepackage{times}
\usepackage{epsfig}
\usepackage{graphicx}
\usepackage{amsmath}
\usepackage{amssymb}

\usepackage[export]{adjustbox}

\makeatletter
\def\thickhline{%
  \noalign{\ifnum0=`}\fi\hrule \@height \thickarrayrulewidth \futurelet
   \reserved@a\@xthickhline}
\def\@xthickhline{\ifx\reserved@a\thickhline
               \vskip\doublerulesep
               \vskip-\thickarrayrulewidth
             \fi
      \ifnum0=`{\fi}}
\makeatother

\newlength{\thickarrayrulewidth}
\usepackage{booktabs,makecell,multirow,tabularx}
\newcolumntype{L}{>{\RaggedRight\arraybackslash}X}
\newcolumntype{C}{>{\Centering\arraybackslash}X}

\newcommand\mc[1]{\multicolumn{3}{c}{#1}}
\usepackage{ragged2e}
\usepackage[accsupp]{axessibility}  

\newcommand{\new}[1]{\textcolor{black}{#1}}

\newcommand{\nikos}[1]{\textcolor{black}{#1}}
\newcommand{\final}[1]{\textcolor{black}{#1}}

\usepackage[pagebackref=true,breaklinks=true,letterpaper=true,colorlinks,bookmarks=false]{hyperref}

\iccvfinalcopy 

\def\iccvPaperID{3796} 

\ificcvfinal\fi

\begin{document}

\title{Bias Loss for Mobile Neural Networks}

\author{Lusine Abrahamyan$^{1,3}$\\

{\tt\small lusine.abrahamyan@vub.be}
\and
Valentin Ziatchin$^{2}$\\
{\tt\small valentin.ziatchin@picsart.com}
\and
Yiming Chen$^{1,3}$\\
{\tt\small cyiming@etrovub.be}
\and
Nikos Deligiannis$^{1,3}$\\
{\tt\small ndeligia@etrovub.be}\\
$^1$ETRO Department, Vrije Universiteit Brussel (VUB), Pleinlaan 2, B-1050 Brussels, Belgium\\
$^2$PicsArt Inc., San Francisco, USA\\
$^3$imec, Kapeldreef 75, B-3001 Leuven, Belgium
}

\maketitle


\begin{abstract}
Compact convolutional neural networks (CNNs) have witnessed exceptional improvements in performance in recent years. However, they still fail to provide the same predictive power as CNNs with a large number of parameters. The diverse and even abundant features captured by the layers is an important characteristic of these successful CNNs. However, differences in this characteristic between large CNNs and their compact counterparts have rarely been investigated.
In compact CNNs, due to the limited number of parameters, abundant features are unlikely to be obtained, and feature diversity becomes an essential characteristic. Diverse features present in the activation maps derived from a data point during model inference may indicate the presence of a set of unique descriptors necessary to distinguish between objects of different classes. In contrast, data points with low feature diversity may not provide a sufficient amount of unique descriptors to make a valid prediction; we refer to them as random predictions. Random predictions can negatively impact the optimization process and harm the final performance. 
This paper proposes addressing the problem raised by random predictions by reshaping the standard cross-entropy to make it biased toward \nikos{data points} with a limited number of unique descriptive features. 
Our novel \textup{Bias Loss} focuses the training on a set of valuable data points and prevents the vast number of samples with poor learning features from misleading the optimization process.
Furthermore, to \nikos{show} the importance of diversity, we present a family of SkipblockNet models whose architectures are brought to boost the number of unique descriptors in the last layers. Experiments conducted on benchmark datasets demonstrate the superiority of the proposed loss function over the cross-entropy loss. Moreover, our SkipblockNet-M can achieve $1\%$ higher classification accuracy than MobileNetV3 Large with similar computational cost on the ImageNet ILSVRC-2012 classification dataset.
The code is available on the link - \url{https://github.com/lusinlu/biasloss_skipblocknet}.
\end{abstract}

\begin{figure}[t]
\begin{center}
   \includegraphics[width=1\linewidth]{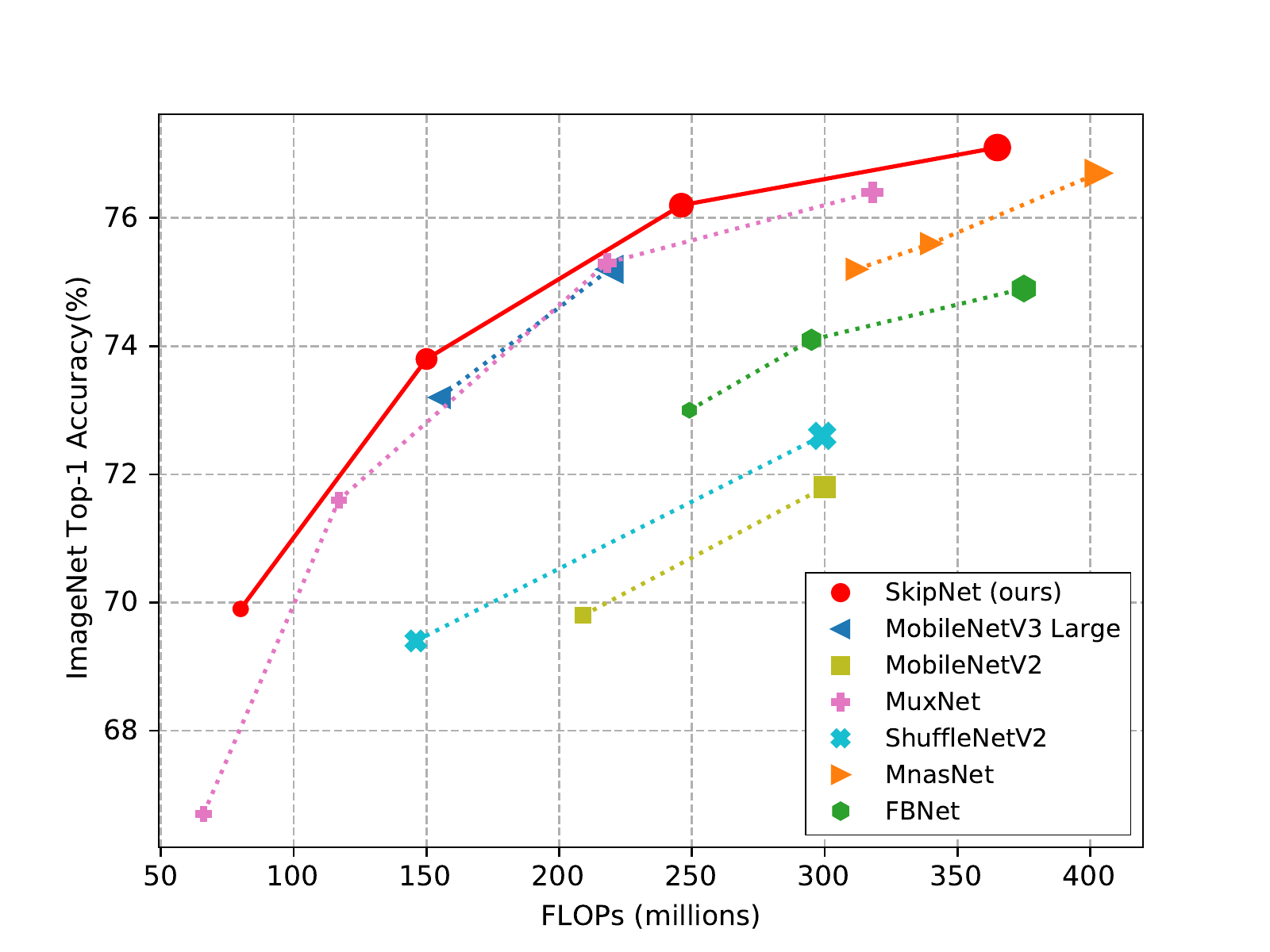}
\end{center}
   \caption{Accuracy~v.s.~FLOPs on ImageNet. \nikos{Our SkipblockNet model} trained with \nikos{the proposed bias loss} outperforms previous well-performing compact neural networks trained with the cross-entropy loss.}
\label{fig:flopsaccuracy}
\end{figure}


\section{Introduction}
Deep CNNs have shown superior performance on numerous computer vision tasks, such as classification, semantic segmentation, and object detection. Typically, models with high predictive power contain a large number of parameters and require a substantial amount of floating point operations (FLOPs); for example, Inception-v3~\cite{inceptionv3} has approximately 24M parameters and requires 6GFLOPs to process an image with a spatial size of $299\times299$~\nikos{pixels}. With the advent of AI applications in mobile devices, several studies have focused on developing high-performance CNNs for resource-constrained settings. \nikos{Several studies have} focused on compressing existing high-performance pretrained models. The compression of the models can be achieved by performing quantization~\cite{wu2016quantized, jacob2018quantization, zhou2016dorefaquantize, rastegari2016xnorquantize, krishnamoorthi2018quantizing, zhou2017incrementalquantize}, pruning~\cite{pruning_3stage, pruning_old, gordon2018morphnetpruning, he2018amcpruning}, or knowledge distillation~\cite{hinton2015distilling, modeldistil}.
Typically, the downside of these methods is an inevitable degradation of performance.

\nikos{Another research line has focused} on designing compact neural networks and architectural units~\cite{lu2020muxconv, han2020ghostnet, tan2019mixconv, zhang2018shufflenet, xception, tan2019efficientnet}.
For example, Xception~\cite{xception} introduced a cost-efficient replacement for the conventional convolution, namely, depthwise separable convolution.
ShuffleNet~\cite{zhang2018shufflenet} replaced convolutional layers with a combination of pointwise group convolution with channel shuffle operation.
\final{The authors of EfficientNet~\cite{tan2019efficientnet} proposed a scaling method that uniformly scales a model's width, depth, and resolution using a set of fixed scaling coefficients.}
\new{However, a significant \nikos{performance improvement} in these methods is mostly connected with \nikos{an} increase in the number of parameters~\cite{tan2019mixconv, han2020ghostnet}. The solution to this problem can be \nikos{the} design of \nikos{a} task-specific objective function. The advantage of designing an objective function over the creation of \nikos{a} new architecture is that \nikos{the former approach} can improve the accuracy of \nikos{a} model without increasing the number of parameters.}
In general, the preferred loss function for classification is \nikos{the} cross-entropy; however, there exist studies indicating that other objectives can outperform the standard cross-entropy loss~\cite{labelsmooth, xie2016disturblabel, focalloss}.
The authors of~\cite{labelsmooth} \nikos{proposed} to compute cross-entropy with the weighted mixture of targets from the uniform distribution. In scenarios where the class imbalance problem exists, \cite{focalloss} proposed to down-weight the loss assigned to well-classified examples. In~\cite{learningweights}, the authors proposed a meta-learning reweighting algorithm in order to tackle the problem of label noise in the dataset.
Although these objectives achieve great performance boost, they target specific problems related mostly to the dataset and do not consider differences between the optimization of compact neural networks and their large counterparts.
Diverse and even abundant information in the feature maps of high-performance CNNs often guarantees a comprehensive understanding of the input data. In compact CNN, due to the small numbers of parameters, \nikos{the} amount of extracted features will be smaller,
and may not be sufficient to describe the object to be classified. For certain data points, these features may lack unique descriptors required to distinguish between the objects of different classes. As a result, in the absence of a sufficient amount of unique descriptors, the model cannot produce a valid prediction. We refer to these as random predictions that contribute no useful learning signal to the optimization process. 

To address this problem, we design \textit{Bias Loss}, a new loss that weights each data point's contribution in proportion to the diversity of features it provides. As a simple measure of diversity, we take the signal's variance, which describes how far the feature maps' values are spread from the average. Based on the variance, we design a nonlinear function, whose values serve as weights for the cross-entropy. This way, we let data points with diverse features have a higher impact on the optimization process and reduce a mislead caused by random predictions.

To further realize bias loss's full potential, we propose the SkipblockNet architecture to address the problem of a lack of extracted features in the last layer. Specifically, we design lightweight intermediate blocks to straightforwardly transfer the low-level features from the first layers to the lasts using skip connections. The usage of the proposed blocks will increase the number of data points with a large number of unique descriptors.
Experimental results showed that the proposed \textit{Bias Loss} is able to boost the performance of the existing mobile models, such as MobileNetV3 Large~\cite{mobilenetv3} ($+0.5\%$), ShuffleNetV2 $0.5\times$~\cite{ma2018shufflenetv2} ($+0.6\%$), SqueezeNet~\cite{iandola2016squeezenet} ($+1\%$). Moreover, SkipblockNet can surpass state-of-the-art compact neural networks such as MobileNetV3, on numerous tasks with fast inference on mobile devices.

To summarize, our contributions are three-fold: (1) we design a loss function to reduce the mislead in the optimization caused by random predictions in compact CNNs; (2) we propose an efficient neural architecture to increase the number of data points with a large number of unique descriptive features; (3) our model achieves state-of-the-art performance on the ImageNet classification task under resource constrained settings.


\section{Related Work}
\new{Many strategies have been proposed for designing compact, computationally efficient, and high-performance CNNs. Bellow, we present two major categories of solutions: the design \nikos{of mobile} architectures and task-oriented objective functions.}
\subsection{Mobile Architectures}

Several CNN architectures have been developed for resource constraint settings~\cite{huang2018condensenet, mobilenets, ma2018shufflenetv2, iandola2016squeezenet, han2020ghostnet, lu2020muxconv}. Among them, the MobileNet~\cite{mobilenets,sandler2018mobilenetv2, mobilenetv3} and ShuffleNet~\cite{zhang2018shufflenet, ma2018shufflenetv2} \nikos{families} stand out due to their high performance achieved with fewer FLOPs. MobileNetV2~\cite{sandler2018mobilenetv2} introduced inverted residual blocks to improve the performance over MobileNetV1~\cite{mobilenets}. Furthermore, MobileNetV3~\cite{mobilenetv3} utilized NAS (Neural Architecture Search) technology~\cite{tan2019mnasnet, dnas_fbnet, dnas_efficient} resulting in achieving higher performance with fewer FLOPs. ShuffleNet~\cite{zhang2018shufflenet} introduced the channel shuffle operation to boost the flow of the information within channel groups. ShuffleNetV2~\cite{ma2018shufflenetv2} further improved the actual speed on hardware.
Despite the high performance achieved with very few FLOPs, the importance of maintaining unique descriptive features in the last layers of the network has never been well exploited.
To that end, we propose SkipblockNet, an architecture that is designed to increase the number of unique descriptive features \nikos{in} the last layers and reduce the number of random predictions. SkipblockNet shares many similarities with the previous high-performance CNNs, in particular, the inverted residual blocks used in MobileNetV3~\cite{mobilenetv3} and the concept of skip connections utilized in U-Net~\cite{unet}. We emphasize that our simple modifications achieve superior results not \nikos{due to} innovation in design but due to the combination of the network with our novel loss. In this way, we can benefit from the developed loss the most.  

\subsection{Objective Functions}

The most common choice for the objective function in many tasks is the cross-entropy. However, various studies have indicated that the design of the loss function, aimed to tackle a specific problem, can have significant benefits~\cite{focalloss, learningweights, labelsmooth, semanticloss, contentloss, huber1992robust}.
Lin \etal~\cite{focalloss} proposed to reshape the standard cross-entropy to address the problem of foreground-background class imbalance encountered during the training of \nikos{an} object detector. The mechanism of label smoothing~\cite{labelsmooth}  suggests using "soft" targets in the cross-entropy calculation. These ``soft'' targets are a weighted mixture of original targets with the uniform distribution over labels. This technique helps preventing the network from becoming over-confident in numerous tasks like image classification, language translation, and speech recognition. Various studies have attempted to address the obstacle caused by noisy labels~\cite{learningweights, labelnoise}. In ~\cite{learningweights} the authors introduce a variation of the weighted cross-entropy, where weights are being learned by the multi-layer perceptron.
The focus of these works has primarily been to optimize the performance of models with a large number of parameters. On the contrary, our loss is designed to tackle the problem arising because of the lack of parameters \nikos{in compact models}, namely the problem of possible mislead in an optimization process caused by random predictions. 

\begin{figure}[t]
\begin{center}
   \includegraphics[width=0.9\linewidth]{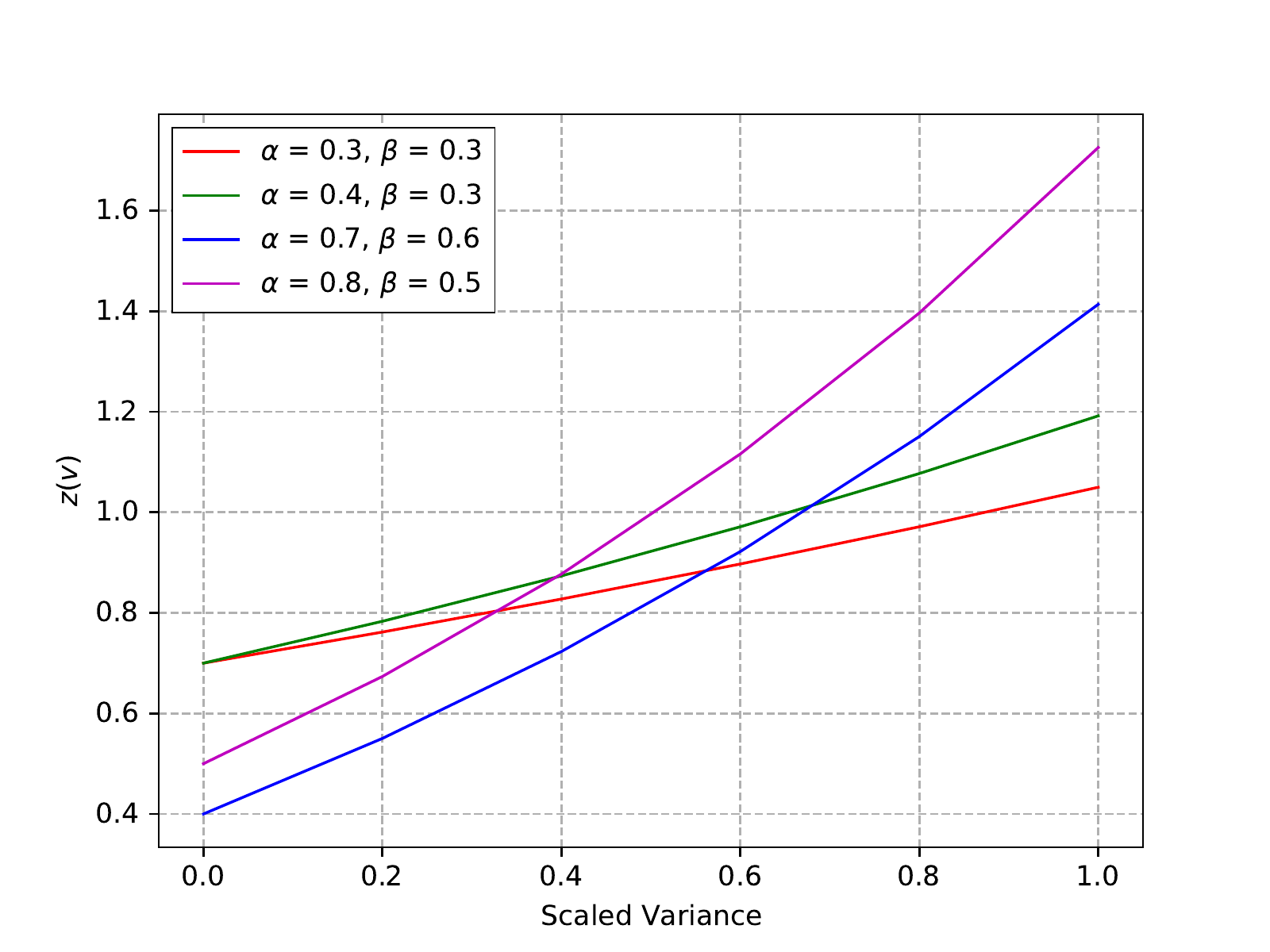}
\end{center}
   \caption{\nikos{The} proposed nonlinear function $z(v)$ given in~\eqref{eq:nonlinearfunction}, where $v$ is \nikos{the} scaled variance. The function comprises two hyperparameters $\alpha$, and $\beta$. An increase of $\beta$ reduces the impact of low variance data points on the cumulative loss. $\alpha$ controls the influence of the high variance data points.}
\label{fig:bias_function}
\end{figure}







\begin{figure}[t]
\begin{center}
   \includegraphics[width=0.9\linewidth]{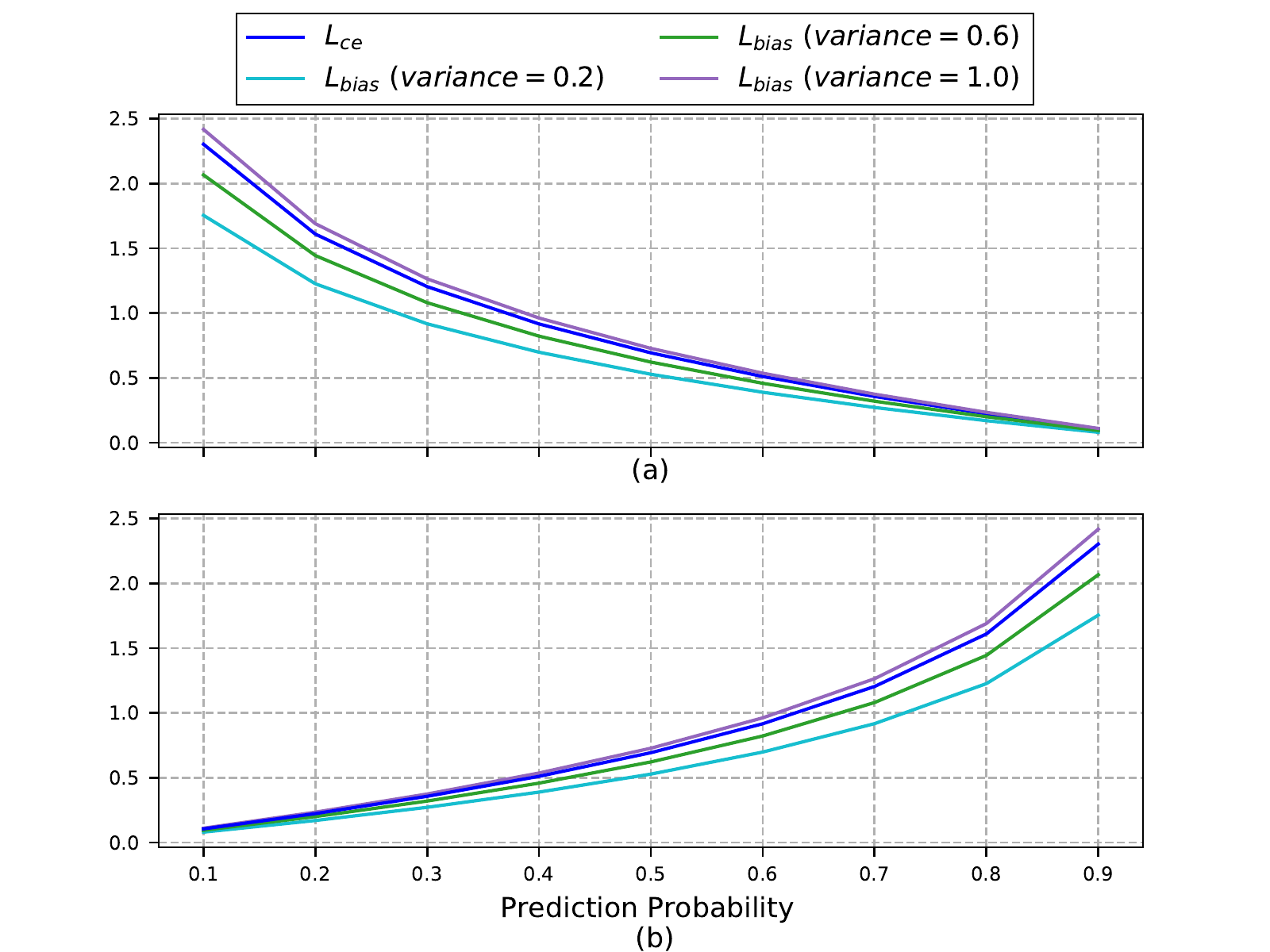}
\end{center}
\caption{Loss v.s. prediction probability (the output of the softmax) for (a) correct predictions and (b) incorrect predictions. \final{$L_{ce}$ denotes the cross-entropy and the $\alpha, \beta$ hyperparameters in the $L_{bias}$ loss are equal to $0.3$.} }
    \label{fig:bias_loss_3d}

\end{figure}

\section{Bias Loss}

We design the \textit{Bias Loss} to address the resource-constrained classification scenario in which there can be a mislead during the optimization process of deep convolutional neural networks~\cite{cnn, vgg, resnet} caused by random predictions.
We advocate that, in compact neural networks, data points \nikos{failing to provide a sufficient amount of unique features that can describe the object} force the model to produce random predictions, that is, predictions made in the absence of feature diversity.

As a simple metric of diversity in all of our experiments, we adopt the signal variance, which can indicate how far the feature maps' values are spread out from the average. The intuition behind \nikos{this} choice is that the higher the variance, the higher the chances of obtaining a large number of unique features. For the variance calculations, the feature maps of the last convolutional layer (before the pooling and dropout operations) are used. This helps avoiding distortions in the results and estimating better the learning signal that \nikos{a} data point provides. 
Let $T \in \mathbb{R}^{b \times c \times h \times w }$ be \nikos{the} output of the convolutional layer, where $b$ is a batch size, $c$ is a number of input channels, and $h$ and $w$ are the height and width of the tensor. Prior to the variance calculations, $T$ is unfolded \nikos{into} a two-dimensional array $t \in \mathbb{R}^{b \times n}$, where $n=c \times h \times w$. The variance of the feature maps of the $i$th data point in the batch is equal to
 \begin{equation}
      v_{i} = \frac{\sum_{j=1}^{n}(t_{j} - \mu)^{2}}{n - 1}, 
\end{equation}
where
\begin{equation}
    \mu = \frac{ \sum_{j=1}^{n}t_{j}} {n}.
\end{equation}
In addition, the variance is scaled to the range $[0, 1]$ for further use in the loss function, that is,
\begin{equation}
    v_{i} = \frac{(v_{i} - min)}{(max - min)},
\end{equation}
where, at each iteration, $max$ and $min$ is the maximum and minimum values of the activations in the batch of feature maps. This is performed to ensure that outliers in the variance values will not lead to large changes in the loss and will not make the model unstable.  

Futhermore, we propose to inject this knowledge about the absence of the unique descriptive features into the optimization process, and to this end, we present the new loss function, namely the \textit{Bias Loss}. 
The bias loss is a dynamically scaled cross-entropy loss, where the scale decays as the variance of data point decreases.

Let $X \in \mathbb{R}^{c \times h \times w }$ be the feature space, where $c$ is a number of input channels and $h$, $w$ are the height and width of the input data, and $Y=\{1, ..., k\}$  be the label space, where $k$ is the number of classes. In a standard scenario, we are given a dataset $D={(x_{i}, y_{i})}_{i=1}^{N}$, where each $(x_{i}, y_{i}) \in X \times Y$, and a neural network $f(x; \theta)$, where $\theta$ denotes the model parameters. Conventionally, the training aims at learning a model by minimizing the expected loss for the training set. In general, the cross-entropy loss for a classification problem is
\begin{equation}
    L_{ce} = - \frac{1}{N} \sum_{i=1}^{N} \sum_{j=1}^{k} y_{ij} \log f_{j}(x_{i}; \theta),
\end{equation}
where we consider that the output layer of the neural network is a softmax.
In order to calibrate the contribution of each data point into the cumulative loss, we propose to add a nonlinear scaling function, which aims at creating a bias between the data points with low and high variance. The bias loss is defined as
\begin{equation}
    L_{bias} = - \frac{1}{N} \sum_{i=1}^{N} \sum_{j=1}^{k} z(v_{i}) y_{ij} \log f_{j}(x_{i}; \theta),
\label{eq:biasloss}
\end{equation}
\begin{equation}
      z(v_{i}) = \exp(v_{i} * \alpha) - \beta,
\label{eq:nonlinearfunction}
\end{equation}
where $\alpha$ and $\beta$ are tunable contribution parameters and $v$ is the scaled variance of the output of the convolutional layer.
The bias function is visualized for several values of $\alpha$ and $\beta$ in Figure~\ref{fig:bias_loss_3d}. We \nikos{notice} two properties of the bias function: (i) when the variance is low, the function values reach their minimum, $(1 - \beta)$,  and the impact of these data points is down-weighted. As the variance increases, the $z(v)$'s values, together with the influence of the data point, exponentially increase.
 (ii) \nikos{The} parameter $\alpha$ smoothly adjusts the rate of the impact of high variance examples. With the increase of $\alpha$, the impact of high variance data points also increases. In addition, Figure~\ref{fig:bias_function} presents the values of the bias loss depending on the variance and the prediction score. The loss is down-weighted mainly for low confidence and low variance data points for both correct and incorrect predictions. Furthermore, it is up-weighted for the high confidence and high variance incorrect predictions, as learning from this kind of data points with a large number of unique features can have a positive impact on the optimization process. 
 Our empirical results suggest that selecting $\alpha=0.3, \beta=0.3$ leads to the best performance.
  
Intuitively, the proposed function helps focusing the learning on examples that can provide a large number of unique features and reducing the possible mislead in the optimization process caused by random predictions.

\section{SkipblockNet Mobile Architectures}
We also introduce a new computational block and a new CNN architecture to further increase the gain in performance obtained via the bias loss. The presented block can be easily integrated into existing architectures and boost the information flow toward the last layers, without additional effort.

\begin{figure}[t]
\begin{center}
  \includegraphics[width=1.0\linewidth]{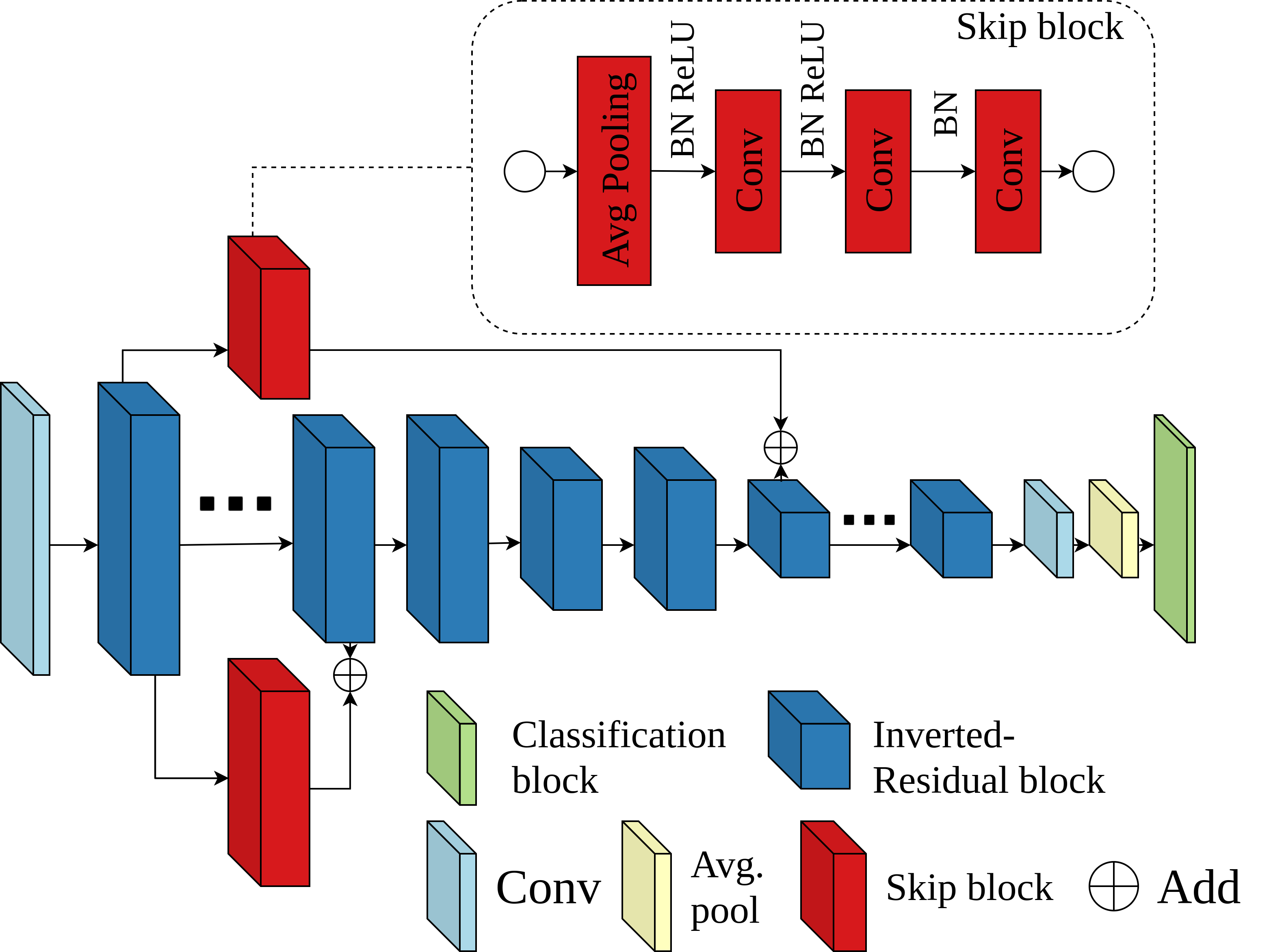}
\end{center}
  \caption{Overview of the SkipblockNet architecture. On top of the inverted residuals~\cite{mobilenetv3}, SkipblockNet uses skip blocks to transfer high-level features of the first block to the last layers. The network design is deliberately simple, which allows concentrating on the contribution of the novel bias loss, which boosts the performance by focusing the training on a set of data points with a rich learning signal.}
\label{fig:SkipblockNet}
\end{figure}


\subsection{Skip Block}

 The idea of the skip block is to deliver the low-level features directly from the first layers to the last ones. The block's design is motivated by the U-Net~\cite{unet} architecture, where, in an autoencoder style architecture, the outputs of layers with the same spatial dimensions in the encoder and decoder are connected via skip connections. Generally, in classification networks, the layers' spatial sizes gradually decrease, making it impossible to use skip connections straightforwardly. To address this limitation, we propose an intermediate block, which is brought to connect layers with different spatial sizes and enrich the last layers with the low-level features extracted from the first layers. As shown in Figure~\ref{fig:SkipblockNet}, the skip block consists of a pooling operation combined with convolutions. First, in order to keep the key features and reduce the spatial sizes, we apply an adaptive average pooling, followed by three convolutional layers. Batch normalization (BN)~\cite{batchnorm} and ReLU nonlinearity~\cite{reluagarap2018deep} are applied after each convolutional layer except for the last one where ReLU is not used. The choice of the adaptive average pooling is motivated by the fact that it takes all features into account, making it possible for the skip block to process all input values. Concerning the convolutional layers' parameters, the setup proposed in MobileNetV3 for the inverted residual blocks is used.

\subsection{SkipblockNet}

Since our primary goal is to boost the number of unique descriptive features in compact neural networks, while mitigating computational complexity, we propose a SkipblockNet architecture \nikos{that} deploys skip blocks.
Due to its superior performance as a design baseline, we follow the architecture of MobileNetV3~\cite{mobilenetv3}. SkipblockNet (\nikos{Figure}~\ref{fig:SkipblockNet}) consists of the stack of inverted residual and the classification blocks of MobileNetV3, and includes our novel skip block. The first layer is a convolution with $16$ filters followed by $15$ inverted residual blocks. Two skip blocks are inserted after the first inverted residual block (\nikos{Figure}~\ref{fig:SkipblockNet}) with the purpose of transferring the information to the sixth and tenth inverted residual blocks. After the skip and inverted residual blocks, a convolutional layer and global average pooling are applied before the final classification block, which consist of dropout and fully connected layers. Similar to MobileNetV3, we use hard-swish nonlinear functions due to their efficiency. As can be seen in Table~\ref{latency}, the latency of SkipblockNet on mobile devices is on par with that \nikos{of} MobileNetV3.
Although the described architecture can already guarantee high performance and low latency, there can be situations where a faster model or higher accuracy may be required. In order to provide a fully customizable network, we integrate the width multiplier, presented in the inverted residual block, into the skip block \nikos{so as to} to control the number of the channels in each layer. By manipulating the width multiplier, the width of the entire network can be changed. That will lead to changes in \nikos{the} model size and computational cost, as well as changes in performance. In general, \nikos{increase of the multiplier will} lead to increase in the performance and latency, and vice versa.
The presented architectures give a basic design for reference, and for further improvement, AutoML methods~\cite{automlbergstra2011algorithms, automlfeurer2015efficient, automlkandasamy2018neural} can be used to tune the skip blocks and boost the performance.


\begin{table*}[t]
\caption{Comparison of state-of-the-art resource constraint neural networks over accuracy, FLOPs, and number of parameters. The results are grouped into sections by FLOPs for better visualization.}
\begin{center}
\scalebox{0.9}{

\begin{tabular}{l||l|l|l|l }
      \thickhline
    Model & FLOPs & Parameters & Top-1 Acc. ($\%$) & Top-5 Acc. ($\%$)\\
      \thickhline
    MobileNetV2 0.5$\times$~\cite{sandler2018mobilenetv2}                 & 97M & 2.0M & 65.4 & 86.4 \\
    MUXNet-xs~\cite{lu2020muxconv}                               & 66M & 1.8M & 66.7 & 86.8 \\
    MobileNetV3 Small 1.0$\times$~\cite{mobilenetv3}           & 66M & 2.9M & 67.4 & - \\
    \textbf{SkipblockNet-XS (with bias loss)}    & 81M & 2.3M & \textbf{69.9} &\textbf{88.9} \\
    \hline
    ShuffleNetV2 $1.0\times$                       & 146M & 2.3M & 69.4 & 88.9 \\
    MUXNet-s~\cite{lu2020muxconv}                           & 117M & 2.4M & 71.6 & 90.3 \\
    ChamNet-C~\cite{dai2019chamnet}             &212M &3.4M & 71.6 & - \\
    MobileNetV3 large $0.75\times$              & 155M    & 4.0M & 73.3 & - \\
    \textbf{SkipblockNet-S (with bias loss)}         & 152M & 3.6M & \textbf{73.8} & \textbf{91.4} \\

    \hline

    FBNet-A~\cite{dnas_fbnet}                                    & 249M & 4.3M & 73.0 & - \\
    MobileNetV3 Large 1.0$\times$               & 219M & 5.4M & 75.2 & - \\
    MUXNet-m                                   & 218M & 3.4M & 75.3 & 92.5\\
    GhostNet 1.3$\times$                         & 226M & 7.3M & 75.7 & 92.7\\
    MixNet-S~\cite{tan2019mixconv}                                    & 256M & 4.1M & 75.8 & 92.8 \\
    \textbf{SkipblockNet-M (with bias loss)}                 & 246M & 5.5M & \textbf{76.2} & \textbf{92.8} \\
    \hline
    ProxylessNAS~\cite{cai2018proxylessnas}                                & 320M & 4.1 & 74.6 & 92.2 \\
    MnasNet-A2~\cite{tan2019mnasnet}                                 & 340M & 4.8M & 75.6 & 92.7\\
    EfficientNet-B0~\cite{tan2019efficientnet}                            & 390M & 5.3 & 76.3 & 93.2 \\
    MUXNet-l                                  & 318M & 4.0M & 76.6 & 93.2\\
    MobileNetV3 large 1.25$\times$            & 356M & 7.5M & 76.6 & - \\
    \textbf{SkipblockNet-L (with bias loss)}   & 364M & 7.1M & \textbf{77.1} & \textbf{93.4} \\
      \thickhline
\end{tabular}
}
\end{center}
\label{imagenettable}
\end{table*}


\section{Experiments}

We present empirical results to demonstrate the efficiency of the novel bias loss and the proposed family of SkipblockNet models. We report results on three tasks: image classification, object detection, and transfer learning. All experiments were performed on a single machine with 2 GeForce RTX 2080 Ti GPUs. \final{Further, during trainings, activation maps with outliers  produce very high variances. In turn, these high variances will lead to  high values of the bias function and make the training unstable. To avoid this effect, in all experiments, we clamp the output of the bias function to the range of [0.5, 1.5].}

\subsection{ImageNet Classification}

We set experiments on ImageNet~\cite{imagenet} and compare \nikos{the} achieved accuracies versus various measures of resource usage such as FLOPs and latency.

\textbf{Training Setup:} ImageNet is a large-scale image classification dataset with over $1.2$M training and $50$K validation images belonging to $1000$ classes.
For experiments on ImageNet, we follow most of the settings used in EfficientNet~\cite{tan2019efficientnet}: \nikos{the} RMSProp optimizer with \nikos{a} decay \nikos{of} $0.9$ and \nikos{a} momentum \nikos{of} $0.9$; \nikos{a} batch norm momentum \nikos{of} $0.99$; \nikos{a} weight decay \nikos{of} $1e-5$; and \nikos{an} initial learning rate \nikos{of} $1e-6$ that increased to $0.032$ in the initial $3$ epochs~\cite{imagenet1hour} and then decays by $0.97$ every $2.4$ epochs. Furthermore, we adopt Inception preprocessing with \nikos{an} image size \nikos{of} $224\times224$~\nikos{pixels}~\cite{inceptionv4}, \nikos{a} batch size of $512$, and complement the training with an exponential moving average with a decay rate of $0.99995$. The reported results are single-crop performance evaluations on the ImageNet validation set.~\new{The aforementioned setting is adopted in order to perform a fair comparison, as most of the state-of-the-art architectures~\cite{tan2019mixconv, han2020ghostnet, lu2020muxconv, mobilenetv3} that we are comparing with are using the same setup.}

\begin{figure}[t]
\begin{center}
   \includegraphics[width=0.9\linewidth]{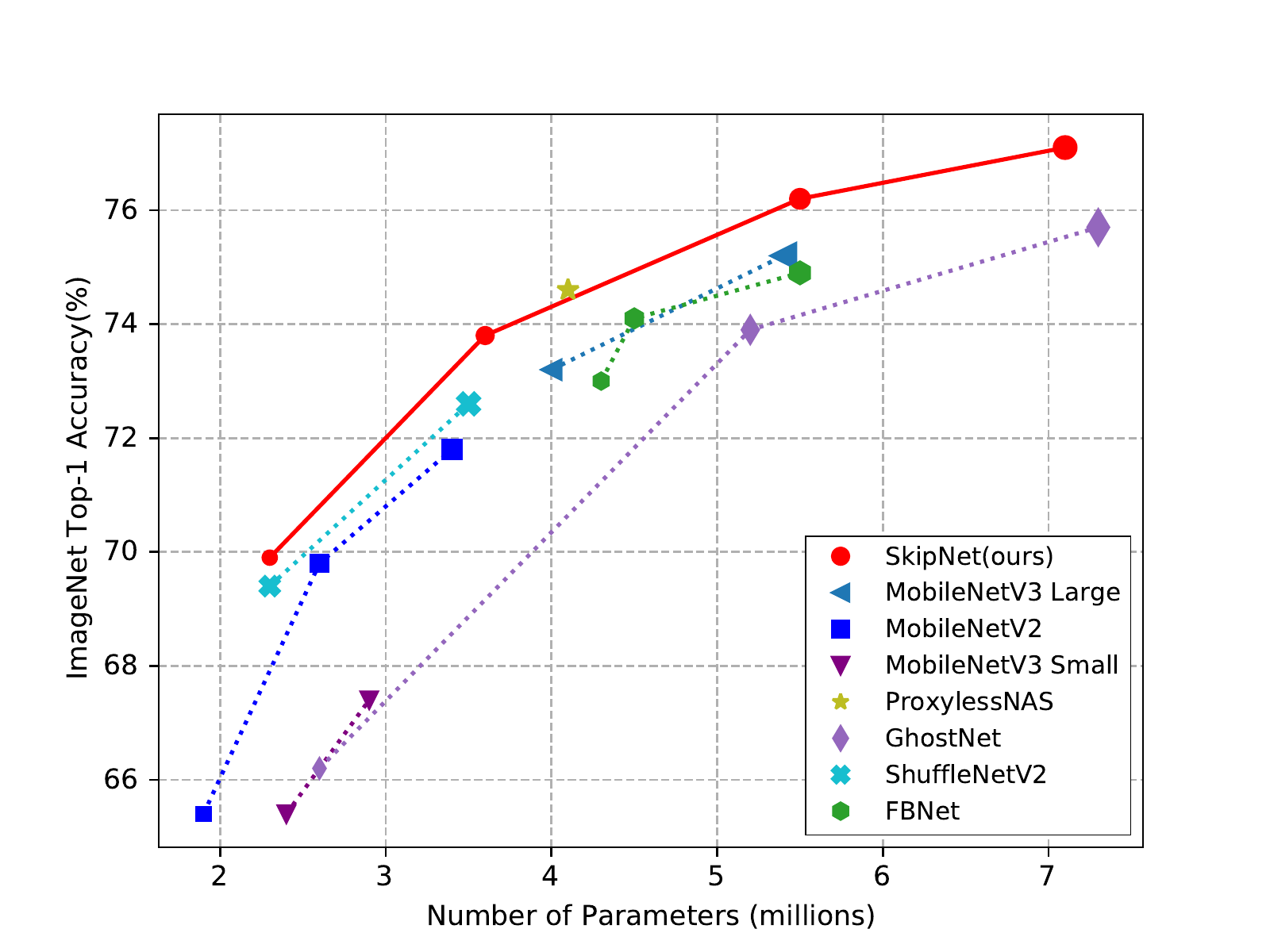}
\end{center}
   \caption{Top-1 classification accuracy v.s. number of parameters on ImageNet for various compact CNNs. Our SkipblockNet is trained with the proposed bias loss.}
\label{fig:paramsaccuracy}
\end{figure}

\textbf{Results:} Table~\ref{imagenettable} shows the performance of the SkipblockNet family models \nikos{in relation to} several modern resource-constraint network architectures. The networks are grouped into four levels of computational complexity: $~50-100$, $~100-200$, ~$200-300$, and ~$300-400$ million FLOPs. We compare them in terms of accuracy, number of parameters and computational complexity (FLOPs). Overall, our family of SkipblockNet models (SkipblockNet-XS, SkipblockNet-S, SkipblockNet-M, SkipblockNet-L) trained with the bias loss outperforms other competitors at the different computational complexity levels. Specifically, SkipblockNet-M archieves $76.2\%$ accuracy with $246$ MFLOPs, which is higher by $1\%$ compared with the MobileNetv3 Large~\cite{mobilenetv3} and by $0.4\%$ compared with MixNet-S~\cite{tan2019mixconv}. Figure~\ref{fig:flopsaccuracy} and Figure~\ref{fig:paramsaccuracy} visualize the trade-off obtained by SkipblockNet and previous compact neural networks.

\begin{table}[t]
\caption{Top-1 accuracy v.s. latency on Google Pixel family phones (Pixel-$n$ denotes a Google Pixel-$n$ phone). All latencies are in ms and are measured using a single core with a batch size of one. \nikos{The top-1} accuracy is calculated on ImageNet.}
\begin{center}
\scalebox{0.85}{

    \begin{tabular}{l||c|c|c }
    \thickhline
     Model &  Top-1 ($\%$) & Pixel 4 & Pixel 3\\
      \thickhline
     SkipblockNet-M                      & 76.2 & 27 & 42  \\
     \hline
     GhostNet $1.3\times$           & 75.7 & 27 & 41  \\
     MnasNet-A2                     & 75.6 & 21 & 39   \\
     MobileNetV3 $1.0\times$        & 75.2 & 26 & 38 \\
     MobileNetV2 $1.0\times$        & 71.8 & 27 & 38 \\
      \thickhline
\end{tabular}}
\end{center}
\label{latency}
\end{table}

\textbf{Inference Speed:} We measure the inference speed of the SkipblockNet-M on Google Pixel phones using the PyTorch V1.6 Mobile framework~\cite{pytorch}. We use a single core in all our measurements. Table~\ref{latency} reports the latencies of the SkipblockNet along with the other state-of-the-art compact neural networks. The results suggest that SkipblockNet can achieve $1\%$ higher accuracy than the MobileNetV3 with computational overhead higher only by $1$ms on Google Pixel 4. 

\begin{table}[t]
\caption{Ablation study for different techniques. \nikos{The baseline} is MobileNetV3 $1.0\times$ and the combination of the baseline with the skip blocks is SkipblockNet-M.}
\begin{center}
\scalebox{0.8}{

    \begin{tabular}{l||ccc}
    \thickhline
      Top-1 ($\%$)      &baseline & skip block & bias loss   \\
      \thickhline
        75.2            & \checkmark & & \\
        75.7            & \checkmark & & \checkmark \\
        75.5            & \checkmark &\checkmark  & \\
        76.2            & \checkmark & \checkmark & \checkmark \\
      \thickhline
\end{tabular}}
\end{center}
\label{techniques}
\end{table}

\new{\textbf{Impact of Different Components on the Performance:}
To investigate the importance of \nikos{the} different techniques used in SkipblockNet, we conduct a series of experiments on the ImageNet dataset, shown in \nikos{Table}~\ref{techniques}.
We \nikos{first consider} the MobileNetV3, a baseline architecture for our SkipblockNet, and trained it with the Bias Loss. As shown in Table~\ref{techniques}, the bias loss can increase the accuracy of MobileNetV3 by $0.5\%$, compared with the training with cross-entropy.
To evaluate the impact of skip blocks, we examined the performance of the baseline MobileNetV3 with the SkipblockNet-M (\nikos{which is the} MobileNetV3 \nikos{architecture} plus skip blocks), both trained with the cross-entropy. \nikos{The results} indicate that \nikos{a} gain of $0.3\%$ can be obtained only by using the skip blocks. Moreover, by enriching the last layers with the low-level information of the first layers, we can increase the number of data points with high variance and make the boost in the performance related to the usage of \nikos{the} bias loss even higher (i.e., \nikos{an} increase of $0.5\%$ in the case of MobileNetV3 and $0.7\%$ for SkipblockNet-M). Moreover, in order to show \nikos{the} skip blocks' advantage over \nikos{a} simple increase of the depth multiplier, we trained MobileNetV3 $1.05\times$ with the 247M FLOPs and 5.9M parameters and compare it with SkipblockNet-M (246M FLOPs, 5.5M parameters). \nikos{When} trained with the cross-entropy, SkipblockNet-M achieves $75.5\%$ accuracy, while MobileNetV3 $1.05\times$ achieves $75.3\%$.}

\subsection{Classification with Bias Loss}

To verify the effectiveness of the proposed bias loss, we apply it on several resource constraint neural networks and conduct experiments using the CIFAR-100~\cite{cifar} classification dataset.
The CIFAR-100 dataset~\cite{cifar} consists of $60,000$ images from $100$ classes.
The dataset is divided into $50,000$ training and $10,000$ testing images.
For training on CIFAR-100, we use an SGD optimizer with a momentum equal to~$0.9$ and \nikos{a} weight decay of $5e-4$. The initial learning rate is set to $1e-1$ and then decays at the epochs $60$, $120$, $160$ at a rate of $0.2$. For data augmentation, images are randomly flipped horizontally and rotated between the angles $[-15, 15]$. \final{ Table~\ref{cifar100} reports the accuracy of neural networks trained with cross-entropy, focal loss~\cite{focalloss} and bias loss. The results shows that models trained with bias loss systematically outperform models trained with cross-entropy and focal loss by about $1\%$ and $0.5\%$, respectively. The results indicate that our loss can boost the performance regardless of the architecture.} In particular, when compared with the cross-entropy, for ShuffleNetV2~\cite{ma2018shufflenetv2} $0.5\times$, the accuracy \nikos{is} increased by $1.5\%$, for SqueezeNet~\cite{iandola2016squeezenet} by $1\%$, and for MobileNetV2 $0.75\times$~\cite{sandler2018mobilenetv2}  by $0.6\%$.

\begin{table}
\caption{Comparison of compact CNNs accuracies trained on CIFAR-100 with the bias loss and cross-entropy.}
\begin{center}
\scalebox{0.8}{

    \begin{tabular}{l||c|c|c|c }
    \thickhline
     Model & Params & Top-1 ($\%$) & Top-1 ($\%$) & Top-1($\%$)\\
           &             & CE loss         & Focal loss & Bias loss \\
      \thickhline
     ShuffleNetV2 $0.5 \times$               & 1.4M & 69.5& 69.8 &\textbf{71} \\
     MobileNetV2 $0.75\times$                 & 2.6M & 68 & 68.2 &\textbf{68.6}\\
     NASNet-A ($N=4$)                          & 5.3M & 77.2& 77.5 &\textbf{78}\\
     SqueezeNet                              & 1.25M & 69.4 & 69.8 &\textbf{70.4} \\
     DenseNet ($k=12$)                        & 7M & 78.9 & 79.5 &\textbf{79.9} \\
      \thickhline
\end{tabular}}
\end{center}
\label{cifar100}
\end{table}

\begin{figure}
\begin{center}
   \includegraphics[width=0.9\linewidth]{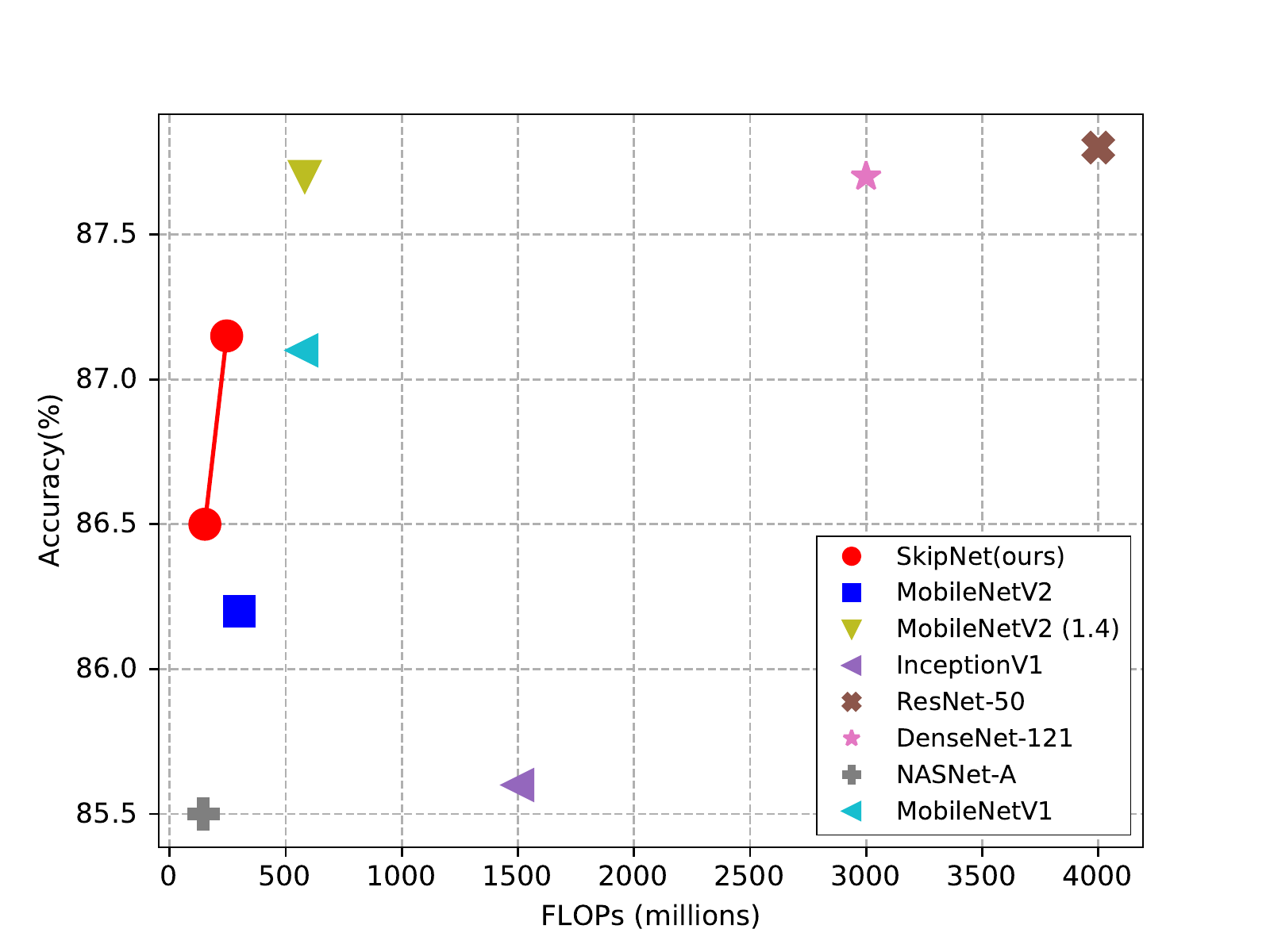}
\end{center}
   \caption{Transfer learning performance. Trade-off between \nikos{top-1} accuracy and number of FLOPs.}
\label{fig:fine_tune}
\end{figure}

\begin{table*}[hbt!]
 \caption{Average/Max/Min variances of the output of the $n$-th and last convolutional layers for different models, where BL and CE indicates trainings with the bias loss and cross-entropy, respectively.}
\begin{center}
\scalebox{0.8}{
   \begin{tabularx}{\textwidth}{@{} l *{9}{C} @{}}
    \toprule
 
\thead{Model}
    &   \mc{\thead{5th layer}} &   \mc{\thead{20th layer}} &   \mc{\thead{last layer}} \\
    \cmidrule(rl){2-4}\cmidrule(rl){5-7}\cmidrule(rl){8-10}
    &  \thead{avg.} &   \thead{max}  &   \thead{min}
    &  \thead{avg.} &   \thead{max}  &   \thead{min}
    &  \thead{avg.} &   \thead{max} &   \thead{min} \\

 \thickhline
     SkipblockNet-M (CE)                &1.7 & 2.4 & 1.6  & 0.6 & 1.2 & 0.1 & 0.09 & 0.2 & 0.04 \\
     SkipblockNet-M (BL)                & \textbf{ 2.}& \textbf{2.5} & \textbf{1.7} & \textbf{1.} & \textbf{1.6}& \textbf{0.2} & \textbf{0.15} & \textbf{0.2} & \textbf{0.09} \\
          \hline
     ShuffleNet (CE)               &1.2 & 1.6 & 0.9  & 0.3 & 0.5 & 0.02 & 0.02 & 0.07 & 0.01   \\
     ShuffleNet (BL)               &\textbf{1.4} & \textbf{1.7} & \textbf{0.9}  & \textbf{0.4} & \textbf{0.7} & \textbf{0.03} & \textbf{0.04} & \textbf{0.1} & \textbf{0.02}   \\
          \hline
     MobileNetV3 $1.0\times$ (CE)  &1.7 & 2.3 & 1.6  &0.4 & 1. & 0.06  &0.05 & 0.09 & 0.01 \\
     MobileNetV3 $1.0\times$ (BL)  &\textbf{1.9} & \textbf{2.4} & \textbf{1.9} &\textbf{0.7} & \textbf{1.5} & \textbf{0.1}  &\textbf{ 0.09} & \textbf{0.1} & \textbf{0.03} \\
     \hline
     Inception V3 (CE)              &3.3 & 5.9 & 1.9  & 5.2 & 9.3 & 2.4 & 0.7 & 3.6 & 0.2  \\
     DenseNet     (CE)              &3 & 6.1 & 1.9  & 4.1 & 7.2 & 1.4  & 0.7 & 2.4 & 0.2 \\
      \thickhline
\end{tabularx}}
\end{center}
\label{variance}
\end{table*}

\begin{table}
\caption{The performance for PASCAL VOC2007 Detection.}
\begin{center}
\scalebox{0.8}{

    \begin{tabular}{l||c|c|c }
    \thickhline
     Model & Parameters & FLOPs & mAP ($\%$)\\
      \thickhline
     VGG + SSD              & 26.2M & 31B & 77.2 \\
     MobileNet + SSD        & 9.4M & 1.6B & 67.5\\
     MobileNetV2 + SSD      & 8.9M & 1.4B & 73.1\\
     \hline
     SkipblockNet-S + SSD        & 9.4M & 1.4B & 73.6 \\
      \thickhline
\end{tabular}}
\end{center}
\label{object_detection}
\end{table}


\subsection{Transfer Learning}

We have also evaluated our SkipblockNet on the transfer learning task using the Food101~\cite{bossard14food} dataset.
Food-101 consists of $75,750$ training and $25,250$ testing images from $101$ different classes.
Figure~\ref{fig:fine_tune} compares the accuracy against FLOPs for our models and the list of other neural networks. Each SkipblockNet model is first trained from scratch on ImageNet and all weights are fine-tuned on the Food101 dataset using a setup similar to~\cite{kornblith2019better}. The accuracy and FLOPs \nikos{results} for the rest of the models are \nikos{taken} from~\cite{kornblith2019better}. \nikos{The} results show that our SkipblockNets significantly outperform previous compact neural networks and have accuracy on par with the models with a large number of parameters. Specifically, SkipblockNet-M achieves  $0.95\%$ higher accuracy, than MobileNetV2~\cite{sandler2018mobilenetv2}, with $1.2\times$ higher efficiency.


\subsection{Object Detection}

To evaluate the generalization ability of SkipblockNet, we conduct object detection experiments on the PASCAL VOC detection benchmark~\cite{pascalvoc}. We use the PASCAL VOC 2012 trainval split as training data and report the mean Average Precision (mAP) on \nikos{the} test split. Our experiments use the Single Shot Detector (SSD)~\cite{ssd} as a detection framework and SkipblockNet as the feature extraction backbone. To set up additional layers, we follow the procedure described in MobileNetV2~\cite{sandler2018mobilenetv2}. We train all the models with the SGD optimizer for $200$ epochs, with \nikos{a} batch size \nikos{of} $42$, \nikos{an} input image size \nikos{of} $300\times300\times3$, \nikos{and an} initial learning rate of $0.01$ with cosine annealing. Table~\ref{object_detection} reports the mAP achieved with the SkipblockNet compared with other models. Under similar resource usage, SkipblockNet-S + SSD achieves $0.5\%$ higher mAP than MobileNetV2~\cite{sandler2018mobilenetv2}.



\subsection{Analysis of the Variance in the Neural Networks}
\final{To show the role of the variance in the CNNs and the impact that the bias loss and the skip blocks can have on it, we conduct experiments on a range of well-known architectures. We examine the distribution of the values in convolutional layers in the networks with a large number of parameters like Inception V3~\cite{inceptionv3} and DensNet169~\cite{huang2017densely} and in compact ones. The purpose of the experiment is to compare the variance in the large and compact models and quantify the boost in the variance that the bias loss and skip blocks can provide.
We took models pre-trained on ImageNet and examined the average, maximum and minimum values of the variance within the different layers.
The results presented in Table 1 indicate that: (1) the variance in large models is significantly higher than that in compact models. Hence, large models can extract a decent amount of descriptive features for almost all samples, and the proposed strategy of reweighting will not boost their performance. (2) The bias loss can increase the variance throughout the model. (3) Skip blocks enrich later layers of a model with low-level features, thereby increasing the variance (SkipblockNet-M (CE) vs MobileNetV3 1.0$\times$(CE)). The increase in the variance leads to the boost in the number of up-weighted data points meaning that, in the case of training with the bias loss, the optimizer will benefit from learning from more useful data points. Hence, the combination of skip blocks with the bias loss can bring a higher gain in accuracy.}

\section{Conclusion}
In this paper, we proposed the \textit{Bias Loss}, a novel loss function designed to improve the performance of the compact CNNs by reducing a mislead during the optimization process caused by the data points with poor learning features. Furthermore,  we presented a family of SkipblockNet models whose architectures are brought to reduce the number of data points with poor learning features. Our extensive experiments, conducted on benchmark datasets and models, illustrate that the proposed loss is able to boost the performance of existing compact CNNs. Moreover, our SkipblockNet-M achieves significantly better accuracy and efficiency than all the latest compact CNNs on the ImageNet classification task.

{\small
\bibliographystyle{ieee_fullname}
\bibliography{egbib}
}

\end{document}